\documentclass[11pt,a4paper]{article}
\usepackage[hyperref]{acl2020}
\usepackage{times}
\usepackage{latexsym}
\usepackage{amsmath}
\usepackage{amssymb}
\usepackage{algorithm, algorithmic, letltxmacro}
\usepackage{graphicx}
\usepackage{multirow}
\usepackage{hyperref}
\usepackage{tikz-dependency}
\usepackage{subfigure}
\usepackage{enumitem}
\usepackage{booktabs}

\usepackage{microtype}

\aclfinalcopy 
\def\aclpaperid{395} 

\title{Efficient Second-Order TreeCRF for Neural Dependency Parsing}

\author{
    Yu Zhang,
    Zhenghua Li\thanks{$~~$Corresponding author},
    Min Zhang \\
    Institute of Artificial Intelligence, School of Computer Science and Technology, \\
    Soochow University, Suzhou, China \\
    {yzhang.cs@outlook.com, \{zhli13,minzhang\}@suda.edu.cn}
}

\date{\today}

\begin{document}
\maketitle

\begin{abstract}
\label{section:abstract}
In the deep learning (DL) era, parsing models are extremely simplified with little hurt on performance, thanks to the remarkable capability of multi-layer BiLSTMs in context representation.
As the most popular graph-based dependency parser
due to its high efficiency and performance, the biaffine parser directly scores single dependencies under the arc-factorization assumption, and adopts a very simple local token-wise cross-entropy training loss.
This paper for the first time presents a second-order TreeCRF extension to the biaffine parser.
For a long time, the complexity and inefficiency of the inside-outside algorithm hinder the popularity of TreeCRF.
To address this issue, we propose an effective way to batchify the inside and Viterbi algorithms for direct large matrix operation on GPUs, and to avoid the complex outside algorithm via efficient back-propagation.
Experiments and analysis on 27 datasets from 13 languages clearly show that techniques developed before the DL era, such as structural learning (global TreeCRF loss) and high-order modeling are still useful, and can further boost parsing performance over the state-of-the-art biaffine parser, especially for partially annotated training data.
We release our code at \url{https://github.com/yzhangcs/crfpar}.
\end{abstract}
\section{Introduction}
\label{section:introduction}

\begin{figure}[tb]
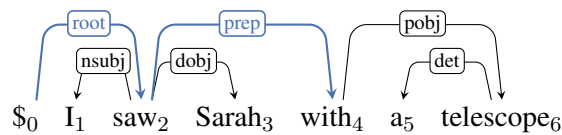

\begin{center}
\begin{dependency}
\begin{deptext}[column sep=.16cm] 
\$$_0$ \& I$_1$ \& saw$_2$ \& Sarah$_3$ \& with$_4$ \&a$_5$ \& telescope$_6$ \\
\end{deptext}
\depedge[edge style={black}]{3}{2}{nsubj}
\depedge[edge style={black}]{3}{4}{dobj}
\depedge[edge style={black}]{5}{7}{pobj}
\depedge[edge style={black}]{7}{6}{det}
\depedge[edge style={draw={rgb,255:red,76; green,114; blue,176}, thick}, label style={draw={rgb,255:red,76; green,114; blue,176}, text={rgb,255:red,76; green,114; blue,176}, semithick}]{1}{3}{root}
\depedge[edge style={draw={rgb,255:red,76; green,114; blue,176}, thick}, label style={draw={rgb,255:red,76; green,114; blue,176}, text={rgb,255:red,76; green,114; blue,176}, semithick}]{3}{5}{prep}
\end{dependency}
\caption{
    An example full dependency tree.
    In the case of partial annotation, only some (not all) dependencies are annotated,
    for example, the two thick (blue) arcs.
}
\label{fig:dep-tree-example} %
\end{center}
\end{figure}

As a fundamental task in NLP, dependency parsing has attracted a lot of research interest 
due to its simplicity and multilingual applicability in capturing both syntactic and semantic information \cite{nivre-lrec2016-UD1.0}.
Given an input sentence $\boldsymbol{x}=w_0w_1 \ldots w_n$,
a dependency tree, as depicted in Figure~\ref{fig:dep-tree-example}, is defined as $\boldsymbol{y}=\{(i,j,l), 0\le i \le n, 1 \le j \le n, l \in \mathcal{L}\}$,
where $(i,j,l)$ is a dependency from the head word $w_i$ to the modifier word $w_j$ with the relation label $l \in \mathcal{L}$.
Between two mainstream approaches, this work focuses on the graph-based paradigm (vs. transition-based).

Before the deep learning (DL) era, graph-based parsing
relies on many hand-crafted 
features
and differs from its neural counterpart in two major aspects.
First, structural learning, i.e., explicit awareness of tree structure constraints during training,
is indispensable. 
Most non-neural graph-based parsers adopt the max-margin training algorithm, which
first predicts a highest-scoring \emph{tree} with the current model,
and then updates feature weights so that the correct tree
has a higher score than the predicted tree.

Second, high-order modeling brings significant accuracy gains. 
The basic first-order model factors the score of a tree into  independent scores of single dependencies 
\cite{mcdonald-etal-2005-online}.
Second-order models were soon propose to incorporate scores of dependency pairs,
such as adjacent-siblings
\cite{mcdonald-pereira-2006-online} and grand-parent-child \cite{carreras-2007-experiments,koo-collins-2010-efficient},
showing significant accuracy improvement yet with the cost of lower efficiency and more complex decoding algorithms.\footnote{Third-order and fourth-order models show little accuracy improvement probably due to the feature sparseness problem \cite{koo-collins-2010-efficient,ma-zhao-2012-fourth}.}


In contrast, neural graph-based 
dependency parsing exhibits an opposite development trend.
\citet{pei-etal-2015-effective} propose to use feed-forward neural networks for automatically learning combinations of dozens of atomic features similar to \citet{chen-manning-2014-fast}, and for computing subtree scores.
They show that incorporating second-order scores of adjacent-sibling subtrees significantly improved performance.
Then, both \citet{wang-chang-2016-graph} and \citet{kiperwasser-goldberg-2016-simple} propose to utilize BiLSTM as an encoder and
use minimal feature sets for scoring single dependencies in a first-order parser.
These three representative works all employ global max-margin training.
\citet{Timothy-d17-biaffine} propose a strong and efficient biaffine parser
and obtain state-of-the-art accuracy on a variety of datasets and languages.
The biaffine parser is also first-order and
employs simpler and more efficient non-structural training
via local head selection for each token \cite{zhang-etal-2017-dependency-parsing}.

Observing such contrasting development, we try to make a connection between pre-DL and DL techniques for graph-based parsing.
Specifically, \textbf{the first question} to be addressed in this work is:
\emph{can previously useful techniques such as structural learning and high-order modeling further improve the state-of-the-art\footnote{
  Though many recent works report higher performance with extra resources,
  for example contextualized word representations learned from large-scale unlabeled texts under language model loss,
  they either adopt the same architecture or achieve similar performance under fair comparison.
}
 biaffine parser,
and if so, in which aspects are they helpful? }

For structural learning, we focus on the more complex and less popular TreeCRF instead of max-margin training.
The reason is two-fold.
First, estimating probability distribution is the core issue in modern data-driven NLP methods \cite{le-zuidema-2014-inside}.
The probability of a tree, i.e., $p(\boldsymbol{y}\mid\boldsymbol{x})$, is potentially more useful than an unbounded score $s (\boldsymbol{x}, \boldsymbol{y})$
for high-level NLP tasks when utilizing parsing outputs.
Second, as a theoretically sound way to measure model confidence of subtrees, 
marginal probabilities can support Minimum Bayes Risk (MBR) decoding \cite{smith-smith-2007-probabilistic}, and are also proven to be crucial
for the important research line of token-level active learning based on partial trees \cite{li-etal-2016-active}.

One probable reason for the less popularity of TreeCRF, despite its usefulness, is
due to the complexity and inefficiency of the inside-outside algorithm, especially the outside algorithm.
As far as we know, all existing works compute the inside and outside algorithms on CPUs.
The inefficiency issue becomes more severe in the DL era, due to the unmatched speed of CPU and GPU computation.
This leads to \textbf{the second question}:
\emph{can we batchify the inside-outside algorithm and perform computation directly on GPUs?}
In that case, we can employ efficient TreeCRF as a built-in component in DL toolkits such as PyTorch
for wider applications \cite{cai-etal-2017-crf,le-zuidema-2014-inside}.







Overall, targeted at the above two questions, this work makes the following contributions.
\begin{itemize}
\item We for the first time propose second-order TreeCRF for neural dependency parsing.
We also propose an efficient and effective triaffine operation for scoring second-order subtrees.
\item We propose to batchify the inside algorithm via direct large tensor computation on GPUs,
leading to very efficient TreeCRF loss computation.
We show that the complex outside algorithm is no longer needed for the computation of gradients and marginal probabilities,
and can be replaced by the equally efficient back-propagation process.
\item We conduct experiments on 27 datasets from 13 languages. The results and analysis show that
both structural learning and high-order modeling are still beneficial to the state-of-the-art biaffine parser in many ways in the DL era.
\end{itemize}


\section{The Basic Biaffine Parser}
\label{section:basic_model}

We re-implement the state-of-the-art biaffine parser \cite{Timothy-d17-biaffine} with
two modifications, i.e., using CharLSTM word representation vectors instead of POS tag embeddings, and
the first-order Eisner algorithm \cite{eisner-2000-iwptbook}
for projective decoding instead of the non-projective MST algorithm.

\begin{figure}[tb]
\centering
\includegraphics{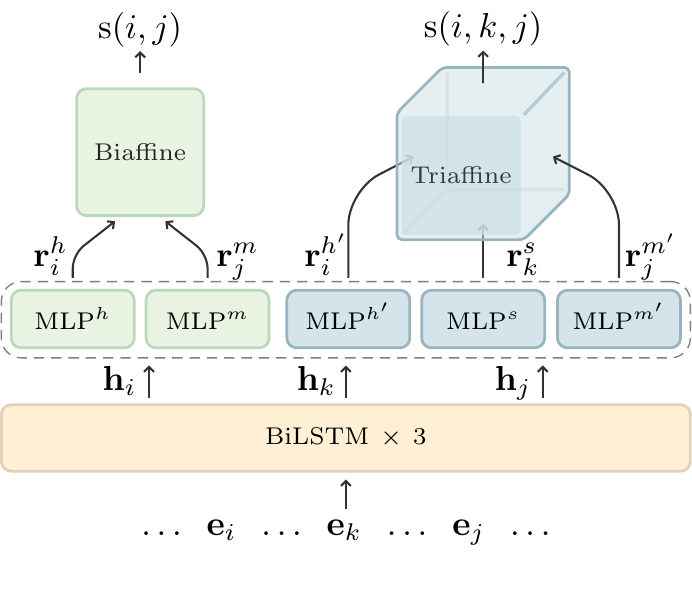}
\caption{Scoring architecture with second-order extension.
}
\label{fig:framework}
\end{figure}

\paragraph{Scoring architecture.}
Figure~\ref{fig:framework} shows the scoring architecture, consisting of four components.





\subparagraph{Input vectors.}
The $i$th input vector 
is composed of two parts:
the word embedding and the CharLSTM word representation vector of $w_i$.
\begin{equation}
\label{equation:input}
\mathbf{e}_i=\mathrm{emb}({w_i}) \oplus \mathrm{CharLSTM}(w_i)
\end{equation}
where $\mathrm{CharLSTM}(w_i)$ is obtained by feeding $w_i$ into a BiLSTM and
then concatenating the two last hidden vectors \cite{lample-etal-2016-neural}.
We find that replacing POS tag embeddings with  $\mathrm{CharLSTM}(w_i)$ leads to consistent improvement,
and also simplifies the multilingual experiments by avoiding POS tag generation (especially n-fold jackknifing on training data).


\subparagraph{BiLSTM encoder.} To encode the sentential contexts,
the parser applies three BiLSTM layers 
over 
$\mathbf{e}_0 \dots \mathbf{e}_n$. The output vector of the top-layer BiLSTM for the $i$th word
is denoted as $\mathbf{h}_i$.

\subparagraph{MLP feature extraction.} Two shared MLPs are applied to $\mathbf{h}_i$, obtaining
two lower-dimensional vectors that detain only syntax-related features:
\begin{equation}
\label{mlp-arc}
\mathbf{r}_i^{h}; \mathbf{r}_i^{m} =\mathrm{MLP}^{h/m} \left( \mathbf{h}_i \right)
\end{equation}
where $\mathbf{r}_i^{h}$ and $\mathbf{r}_i^{m}$ are the representation vector of $w_i$ as a head word and a modifier word respectively.

\subparagraph{Biaffine scorer.}
\citet{Timothy-d17-biaffine} for the first time propose to compute the score of a dependency $i \rightarrow j$ via biaffine attention:
\begin{equation} \label{equation:biaffine}
s(i,j) =  \left[
\begin{array}{c}
  \mathbf{r}_{j}^{m}    \\
    1
\end{array}
\right]^\mathrm{T}
\mathbf{W}^\textit{biaffine}  \mathbf{r}_{i}^{h}
\end{equation}
where $\mathbf{W}^\textit{biaffine} \in \mathbb{R}^{d \times d}$.
The computation is 
extremely efficient on GPUs.

\paragraph{Local token-wise training loss.}
The biaffine parser adopts a simple non-structural training loss,
trying to independently maximize the local probability of the correct head word for each word.
For a gold-standard head-modifier pair ($w_i$, $w_j$)
in a training instance,
the cross-entropy loss is
\begin{equation} \label{equation:biaffine-loss}
\mathit{L}(i,j) = -\log{\frac{e^{s(i,j)}}{\sum_{0 \le k \le n} e^{s(k,j)}}}
\end{equation}
In other words, the model is trained based on simple head selection,
without considering the tree structure at all, and
losses of all words in a mini-batch are accumulated.

\paragraph{Decoding.} Having scores of all dependencies,
we adopt the first-order Eisner algorithm with time complexity of $O(n^3)$
to find the optimal tree.
\begin{equation}
\label{equation:map-decoding}
{\boldsymbol{y}}^* = \arg\max_{\boldsymbol{y}} \left[ s(\boldsymbol{x},\boldsymbol{y}) \equiv
\sum_{i \rightarrow j \in \boldsymbol{y}}{s(i,j)} \right]
\end{equation}

\paragraph{Handling dependency labels.}
The biaffine parser treats skeletal tree searching and labeling as two independent (training phase) and cascaded (parsing phase) tasks.
This work follows the same strategy for simplicity. Please refer to \citet{Timothy-d17-biaffine} for details.

\section{Second-order TreeCRF}
\label{2o-tree-crf}


This work substantially extends the biaffine parser in two closely related aspects:
using probabilistic TreeCRF for structural training and explicitly incorporating high-order subtree scores.
Specifically, we further incorporate adjacent-sibling subtree scores into the basic first-order model:\footnote{
  This work can be further extended to incorporate
  grand-parent-modifier subtree scores
  based on the viterbi algorithm of $O(n^4)$ time complexity proposed by \citet{koo-collins-2010-efficient}, which
  we leave for future work.
}
\begin{equation}\label{eq:score-definition-2o}
s(\boldsymbol{x}, \boldsymbol{y}) = \sum_{i\rightarrow j \in \boldsymbol{y}}s(i,j) + \sum_{
  i\rightarrow \{k,j\} \in \boldsymbol{y} 
} s(i,k,j)
\end{equation}
where $k$ and $j$ are two adjacent modifiers of $i$ and satisfy either $i < k < j$ or $j < k < i$.

As a probabilistic model, TreeCRF computes the conditional probability of a tree as
\begin{equation}\label{equation:prob-labeled}
\begin{split}
& p(\boldsymbol{y}\mid\boldsymbol{x})  = \frac{e^{s(\boldsymbol{x},\boldsymbol{y})}}{Z(\boldsymbol{x}) \equiv \sum_{\boldsymbol{y'} \in \mathcal{Y}(\boldsymbol{x})} {e^{s(\boldsymbol{x},\boldsymbol{y'})}}}
\end{split}
\end{equation}
where $\mathcal{Y}(\boldsymbol{x})$ is the set of all legal (projective) trees for $\boldsymbol{x}$, and
$Z(\boldsymbol{x})$ is commonly referred to as the normalization (or partition) term.

During training, TreeCRF employs the following structural training loss to
maximize the conditional probability of the gold-standard tree $\boldsymbol{y}$ given $\boldsymbol{x}$.
\begin{equation}\label{equation:training-loss-treecrf}
\begin{split}
\mathit{L}(\boldsymbol{x},\boldsymbol{y}) &= -\log p(\boldsymbol{y}\mid\boldsymbol{x})  \\
&= - s(\boldsymbol{x}, \boldsymbol{y}) + \log Z(\boldsymbol{x})
\end{split}
\end{equation}

\begin{figure}[tb]
\centering
\subfigure {
  \centering
  \begin{minipage}{.97\columnwidth}
  \includegraphics{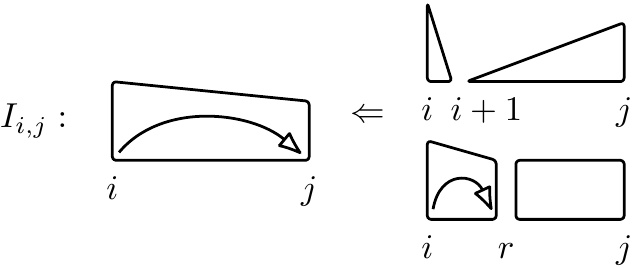}
  \label{fig:eisner-2o-a}
  \end{minipage}
}
\subfigure {
  \centering
  \begin{minipage}{.97\columnwidth}
  \includegraphics{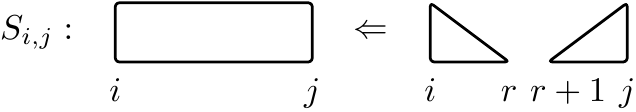}
  \label{fig:eisner-2o-b}
  \end{minipage}
}
\subfigure {
  \begin{minipage}{.97\columnwidth}
  \includegraphics{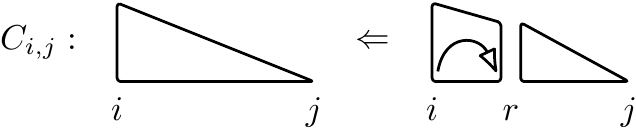}
  \label{fig:eisner-2o-c}
  \end{minipage}
}
\caption{Diagrams of the second-order inside 
algorithm based on bottom-up dynamic programming.}
\label{fig:eisner-2o}
\end{figure}

\subsection{Scoring Second-order Subtrees}
To avoid major modification to the original scoring architecture,
we take a straightforward extension to obtain scores of adjacent-sibling subtrees.
First, we employ three extra MLPs to perform similar feature extraction.
\begin{equation}
\label{mlp-sib}
\mathbf{r}_i^{h'}; \mathbf{r}_i^{s}; \mathbf{r}_i^{m'} =\mathrm{MLP}^{h'/s/m'} \left( \mathbf{h}_i \right)
\end{equation}
where $\mathbf{r}_i^{h'}; \mathbf{r}_i^{s}; \mathbf{r}_i^{m'}$ are the representation vectors of $w_i$ as
head, sibling, and modifier respectively.\footnote{
 Another way is to use one extra MLP for sibling representation, and re-use head and modifier representation from the basic first-order components, which however leads to inferior performance
   in our preliminary experiments.
}


Then, we
propose a natural extension to the biaffine equation, and employ triaffine for score computation over three vectors.\footnote{
  We have also tried the approximate method of \citet{wang-etal-2019-second}, which uses three
  biaffine operations to simulate the interactions of three input vectors, but observed inferior performance.
  We omit the results due to the space limitation.
}
\begin{equation} \label{equation:triaffine}
s(i,k,j) =
\left[
\begin{array}{c}
\mathbf{r}_{k}^{s} \\
1
\end{array}
\right]^\mathrm{T}
{\mathbf{r}_{i}^{h'}}^\mathrm{T}
\mathbf{W}^\textit{triaffine}
\left[
\begin{array}{c}
\mathbf{r}_{j}^{m'} \\
1
\end{array}
\right]
\end{equation}
where $\mathbf{W}^\textit{triaffine} \in \mathbb{R}^{d' \times d' \times d'}$ is a three-way tensor.
The triaffine computation can be quite efficiently performed with the $\mathrm{einsum}$ function on PyTorch.

\subsection{Computing TreeCRF Loss Efficiently}


The key to TreeCRF loss is how to efficiently compute $\log Z(\boldsymbol{x})$,
as shown in Equation~\ref{equation:training-loss-treecrf}.
This problem has been well solved long before the DL era
for non-neural dependency parsing.
Straightforwardly, we can directly extend the viterbi decoding algorithm by replacing max product with sum product, and naturally obtain $\log Z(\boldsymbol{x})$ in the same polynomial time complexity.
However, it is not enough to solely perform the inside algorithm for non-neural parsing, due to the inapplicability of the automatic differentiation mechanism. 
In order to obtain marginal probabilities and then feature weight gradients, we have to realize the more sophisticated outside algorithm, which is usually at least twice slower than the inside algorithm.
This may be the major reason for the less popularity of TreeCRF (vs. max-margin training) before the DL era.

\begin{algorithm}[tb]
\begin{algorithmic}[1]
\newlength{\commentindent}
\setlength{\commentindent}{.2\textwidth}
\renewcommand{\algorithmiccomment}[1]{\unskip\hfill\makebox[\commentindent][l]{$\rhd$~#1}\par}
\LetLtxMacro{\oldalgorithmic}{\algorithmic}
\renewcommand{\algorithmic}[1][0]{%
  \oldalgorithmic[#1]%
  \renewcommand{\ALC@com}[1]{%
  \ifnum\pdfstrcmp{##1}{default}=0\else\algorithmiccomment{##1}\fi}%
}
\begin{footnotesize}
\STATE \textbf{define:} $I,S,C \in \mathbb{R}^{n \times n \times B}$ \COMMENT{$B$ is \#sents in a batch}
\STATE \textbf{initialize:} $C_{i, i} = \log e^0 = 0, 0 \le i \le n$

\FOR [span width]{$w = 1$ \TO $n$}
  \STATE \textbf{Batchify:} $0 \le i$; $j=i+w \le n$
  \STATE
    $I_{i, j} = \log\left(
      \begin{array}{l}
        e^{C_{i, i}  +  C_{j, i+1}} ~ + \\
      \sum\limits_{i < r < j} e^{I_{i, r} + S_{r, j}
      + s(i, r, j)} \\

      \end{array}
      \right)$ + s(i, j)


  \STATE $S_{i, j} = \log \sum\limits_{i \le r < j} e^{C_{i, r}  +  C_{j, r+1}} $ \\
  \STATE $C_{i, j} = \log
    \sum\limits_{i < r \le j} e^{I_{i, r}  +  C_{r, j}}  $ \\
\ENDFOR \COMMENT{refer to Figure~\ref{fig:eisner-2o}}
\RETURN $C_{0, n} \equiv \log Z$
\end{footnotesize}
\end{algorithmic}
\caption{Second-order Inside Algorithm.}
\label{alg:eisner-2o}
\end{algorithm}

As far as we know, all previous works on neural TreeCRF parsing
explicitly implement the inside-outside algorithm for gradient computation \cite{zhang-etal-2019-empirical, jiang-etal-2018-supervised}.
To improve efficiency, computation is transferred from GPUs to CPUs with Cython programming.

This work shows that the inside algorithm
can be effectively batchified to fully utilize the power of GPUs.
Figure~\ref{fig:eisner-2o} and Algorithm \ref{alg:eisner-2o} together illustrate the batchified version of the second-order inside algorithm, which is a direct extension of the second-order Eisner algorithm in \citet{mcdonald-pereira-2006-online} by replacing max product with sum product.
We omit the generations of incomplete, complete, and sibling spans in the opposite direction from $j$ to $i$ for brevity.



Basically, we first pack the scores of same-width spans at different positions
($i, j$) for all $B$ sentences in the data batch into large tensors.
Then we can do computation and aggregation simultaneously on GPUs via efficient large tensor operation.

Similarly, we also batchify the decoding algorithm. Due to space limitation, we omit the details.

It is noteworthy that the techniques described here are also applicable to other grammar formulations such as CKY-style constituency parsing \cite{finkel-etal-2008-efficient,drozdov-etal-2019-unsupervised-latent}.

\subsection{Outside via Back-propagation}

\citet{eisner-2016-inside} proposes a theoretical proof on the equivalence between
the back-propagation mechanism and the outside algorithm in the case of constituency (phrase-structure) parsing.
This work empirically verifies this equivalence for dependency parsing.

Moreover, we also find that marginal probabilities $p(i \rightarrow j\mid\boldsymbol{x})$ directly correspond to gradients after back-propagation with $\log Z(\boldsymbol{x})$ as the loss:
\begin{equation}
\label{equation:partial-derivative}
\begin{split}
\frac{\partial \log Z}{\partial \mathrm{s}(i, j)} 
&= \sum_{\boldsymbol{y}:(i,j) \in \boldsymbol{{y}}} p(\boldsymbol{y}\mid\boldsymbol{x}) = p(i \rightarrow j\mid\boldsymbol{x})
\end{split}
\end{equation}
which can be easily proved. 
For TreeCRF parsers, we perform MBR decoding \cite{smith-smith-2007-probabilistic} by replacing scores with marginal probabilities in the decoding algorithm,
leading to a slight but consistent accuracy increase.

\subsection{Handling Partial Annotation}
\label{sub@section:partial-annotation}

As an attractive research direction, studies show that
 it is more effective to construct or even collect partially labeled data \cite{nivre-etal-2014-squibs,hwa-99-partial-annotation,pereira-92-inside-outside},
 where 
a sentence may correspond to a partial tree $|{\boldsymbol{y}^p}| < n$ in the case of dependency parsing.
Partial annotation can be very powerful when combined with active learning, because
annotation cost can be greatly reduced if annotators only need to annotate sub-structures that are difficult for models. 
\citet{li-etal-2016-active} present a detailed survey on this topic.
Moreover, \citet{peng2019overview} recently released a partially labeled multi-domain Chinese dependency treebank based on this idea.

Then, the question is how to train models on partially labeled data.
\citet{li-etal-2016-active} propose to extend TreeCRF for this purpose and obtain promising results
in the case of non-neural dependency parsing.
This work applies their approach to the neural biaffine parser.
We are particularly concerned at the influence of structural learning and high-order modeling on the utilization of partially labeled training data.

For the basic biaffine parser based on first-order local training, it seems  the only choice is omitting losses of unannotated words.
In contrast, tree constraints allow annotated dependencies to influence the
probability distributions of unannotated words, and high-order modeling further helps by promoting inter-token interaction.
Therefore, both structural learning and high-order modeling are intuitively very beneficial.

Under partial annotation, we follow \citet{li-etal-2016-active} and define the training loss as:
\begin{equation}
\label{equation:training-loss-treecrf-partial}
\begin{split}
\mathit{L}(\boldsymbol{x}, {\boldsymbol{y}^p}) &= -\log \sum\limits_{\boldsymbol{y} \in \mathcal{Y}(\boldsymbol{x}); \boldsymbol{y} \supseteq {\boldsymbol{y}^p}} p(\boldsymbol{y}\mid\boldsymbol{x})  \\
&= - \log \frac{Z(\boldsymbol{x}, {\boldsymbol{y}^p}) \equiv \sum\limits_{\boldsymbol{y} \in \mathcal{Y}(\boldsymbol{x}); \boldsymbol{y} \supseteq \boldsymbol{y}^p} e^{s(\boldsymbol{x},\boldsymbol{y})}}{Z(\boldsymbol{x})}
\end{split}
\end{equation}
where $Z(\boldsymbol{x}, {\boldsymbol{y}^p})$ only considers all legal trees that are compatible with the given partial tree and can also be efficiently computed like $Z(\boldsymbol{x})$.









\section{Experiments}
\label{section:experiments-analysis}

\paragraph{Data.}
We conduct experiments and analysis on 27 datasets from 13 languages,
including two widely used datasets:
the English Penn Treebank (PTB) data with Stanford dependencies \cite{chen-manning-2014-fast}, 
and the Chinese data at the CoNLL09 shared task \cite{hajic-etal-2009-conll}.

We also adopt the Chinese dataset released at the NLPCC19 cross-domain dependency parsing shared task \cite{peng2019overview},
containing one source domain and three target domains.
For simplicity,
we directly merge the train/dev/test data of the four domains into larger ones respectively.
One characteristic of the data is that most sentences are partially annotated based on active learning.

Finally, we conduct experiments on Universal Dependencies (UD) v2.2 and v2.3
following \citet{ji-etal-2019-graph} and \citet{zhang-etal-2019-empirical} respectively.
We adopt the 300d multilingual pretrained word embeddings used in \citet{zeman-etal-2018-conll} and take the CharLSTM representations as input.
For UD2.2, to compare with \citet{ji-etal-2019-graph}, we follow the raw text setting of the CoNLL18 shared task \cite{zeman-etal-2018-conll}, and directly use their sentence segmentation and tokenization results.
For UD2.3, we also report the results of using gold-standard POS tags to compare with \citet{zhang-etal-2019-empirical}.




\paragraph{Evaluation metrics.}
We use unlabeled and labeled 
attachment score (UAS/LAS) 
as the main metrics.
Punctuations are omitted for PTB.
For the partially labeled NLPCC19 data, we adopt the official evaluation script, which simply omits the words without gold-standard heads to accommodate partial annotation.
We adopt Dan Bikel’s randomized parsing evaluation comparator for significance
test.

\paragraph{Parameter settings.}
We directly adopt most parameter settings of \citet{Timothy-d17-biaffine}, including dropout and initialization strategies.
For CharLSTM, the dimension of input char embeddings is 50, and the dimension of output vector is 100, following \citet{lample-etal-2016-neural}.
For the second-order model,
we set the dimensions of $\mathbf{r}^{h'/s/m'}_i$ to 100,
and find little accuracy improvement when increasing to 300.
%
%
We trained each model for at most 1,000 iterations, and stop training if the peak performance on the dev data
does not increase in 100 consecutive epochs.

\paragraph{Models.}
\textsc{Loc} uses local cross-entropy training loss and employs the Eisner algorithm for finding the optimal projective tree.
\textsc{Crf} and \textsc{Crf2o} denote the first-order and second-order TreeCRF model respectively.
$\textsc{Loc}_{\textsc{mst}}$ denotes the basic local model that directly produces non-projective tree based on the MST decoding algorithm of \citet{Timothy-d17-biaffine}.



\subsection{Efficiency Comparison}



\begin{figure}[tb]
\centering
\includegraphics{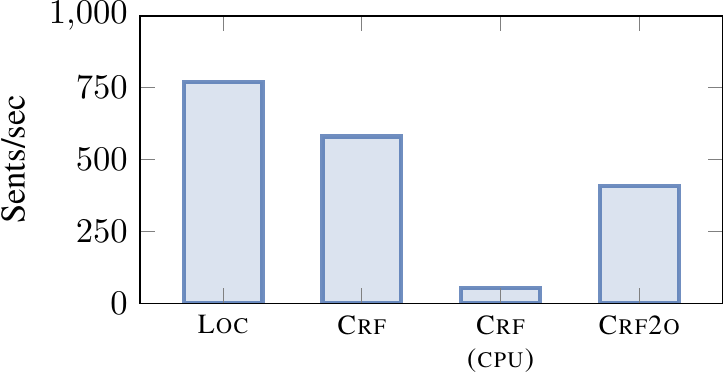}
\caption{
    Parsing speed comparison on PTB-test.
}
\label{fig:speed}
\end{figure}

Figure~\ref{fig:speed} compares the parsing speed of different models on PTB-test.
For a fair comparison, we run all models on the same machine with Intel Xeon CPU (E5-2650v4, 2.20GHz) and GeForce GTX 1080 Ti GPU.
``\textsc{Crf (cpu)}'' refers to the model that explicitly
performs the inside-outside algorithm using Cython on CPUs.
Multi-threading is employed since sentences are mutually independent.
However, we find that using more than 4 threads does not further improve the speed.

We can see that the efficiency of TreeCRF is greatly improved by batchifying the inside algorithm and implicitly
realizing the outside algorithm by back-propagation on GPUs.
For the first-order \textsc{Crf} model,
our implementation can parse about 500 sentences per second, over 10 times faster than the multi-thread ``\textsc{Crf (cpu)}''.
For the second-order \textsc{Crf2o}, our parser achieves the speed of 400 sentences per second, which is able to meet the requirements of a real-time system.
More discussions on efficiency are presented in Appendix~\ref{section:decoding-efficiency}.




\begin{table}[tb]
\setlength{\tabcolsep}{4.2pt}
\centering
\begin{tabular}{lcccc}
\toprule
& \multicolumn{2}{c}{Dev} & \multicolumn{2}{c}{Test} \\
& UAS & LAS & UAS & LAS \\[2pt]
\hline
\\[-8pt]
\multicolumn{5}{c}{PTB} \\
Biaffine17 & - & - & 95.74 & 94.08 \\
F\&K19 & - &  -  & -     & 91.59 \\
Li19 & 95.76 &  93.97 & 95.93 & 94.19 \\
Ji19 & 95.88 & 93.94 & 95.97 & 94.31 \\
Zhang19 & - & - & -  & 93.96 \\[3pt]
\textsc{Loc}             & 95.82          & 93.99          & 96.08          & 94.47 \\
\textsc{Crf} w/o MBR     & 95.74          & 93.96          & 96.04          & 94.34 \\
\textsc{Crf}             & 95.76          & 93.99          & 96.02          & 94.33 \\
\textsc{Crf2o} w/o MBR   & \textbf{95.92} & \textbf{94.16} & \textbf{96.14} & \textbf{94.49} \\
\textsc{Crf2o}           & 95.90          & 94.12          & 96.11          & 94.46 \\[2pt]
\hline
\\[-8pt]
\multicolumn{5}{c}{CoNLL09} \\
Biaffine17 & - & - & 88.90 & 85.38 \\
Li19 & 88.68 & 85.47 & 88.77 & 85.58 \\[3pt]
\textsc{Loc} & 89.07 & 86.10 & 89.15 & 85.98 \\
\textsc{Crf} w/o MBR   &         89.04  &         86.04  &         89.14                    &         86.06 \\
\textsc{Crf}           &         89.12  &         86.12  &         89.28                    &         86.18\rlap{$^\dagger$} \\
\textsc{Crf2o} w/o MBR &         89.29  &         86.24  &         89.49                    &         86.39 \\
\textsc{Crf2o}         & \textbf{89.44} & \textbf{86.37} & \textbf{89.63}\rlap{$^\ddagger$} & \textbf{86.52}\rlap{$^\ddagger$} \\[2pt]
\hline
\\[-8pt]
\multicolumn{5}{c}{NLPCC19} \\
\textsc{Loc}           &         77.01  &         71.14  &         76.92                    &         71.04 \\
\textsc{Crf} w/o MBR   &         77.40  &         71.65  &         77.17                    &         71.58 \\
\textsc{Crf}           &         77.34  &         71.62  &         77.53\rlap{$^\ddagger$}  &         71.89\rlap{$^\ddagger$} \\
\textsc{Crf2o} w/o MBR &         77.58  &         71.92  &         77.89                    &         72.25 \\
\textsc{Crf2o}         & \textbf{78.08} & \textbf{72.32} & \textbf{78.02}\rlap{$^\ddagger$} & \textbf{72.33}\rlap{$^\ddagger$} \\
\bottomrule
\end{tabular}
\caption{Main results. We perform significance test against \textsc{Loc} on the test data, where ``$\dagger$'' means $\mathrm{p} < 0.05$ and ``$\ddagger$'' means $\mathrm{p} < 0.005$.
Biaffine17: \citet{Timothy-d17-biaffine}; F\&K19: \citet{falenska-kuhn-2019-non};
Li19: \citet{li-etal-2019-attentive}; Ji19: \citet{ji-etal-2019-graph};
Zhang19: \citet{zhang-etal-2019-empirical}.
}
\label{table:dev-test}
\end{table}

\begin{figure*}[tb]
\centering
\includegraphics[width=\textwidth]{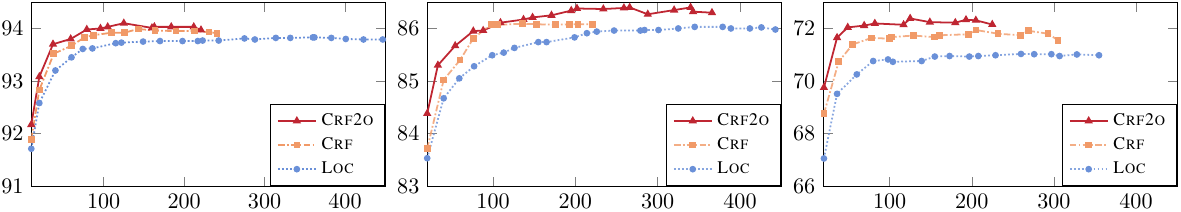}
\caption{
    Convergence curves (LAS vs. training epochs) on dev data of PTB, CoNLL09, and NLPCC19.
}
\label{fig:convergency}
\end{figure*}

\subsection{Main Results}


Table~\ref{table:dev-test} lists the main results on the dev and test data.
The trends on dev and test are mostly consistent.
For a fair comparison with previous works, we only consider those without using extra resources such as ELMo \cite{peters-etal-2018-deep} and BERT \cite{devlin-etal-2019-bert}.
We can see that our baseline \textsc{Loc} achieves the best performance on both PTB and CoNLL09.


On PTB, both \textsc{Crf} and \textsc{Crf2o} fail to improve the parsing accuracy further,
probably because the performance is already very high.
However, as shown by further analysis in Section \ref{section:analysis},
the positive effect is actually introduced by structural learning and high-order modeling.

On CoNLL09, \textsc{Crf} significantly outperforms \textsc{Loc},
and \textsc{Crf2o} can further improve the performance.

On the partially annotated NLPCC19 data, \textsc{Crf} outperforms \textsc{Loc} by a very large margin,
indicating the usefulness of structural learning in the scenario of partial annotation.
\textsc{Crf2o} further improves the parsing performance by explicitly modeling second-order subtree features.
These results confirm our intuitions discussed in Section \ref{sub@section:partial-annotation}.
Please note that the parsing accuracy looks very low because the partially annotated tokens
are usually difficult for models.






\subsection{Analysis}
\label{section:analysis}


\begin{table}[tb]
\setlength{\tabcolsep}{5pt}
\centering
\begin{tabular}{lccccc}
\toprule
& \multicolumn{3}{c}{SIB} & \multirow{2}{*}{UCM} & \multirow{2}{*}{LCM} \\
& P & R & F \\[2pt]
\hline
\\[-9pt]
\multicolumn{6}{c}{PTB} \\
\textsc{Loc}  & 91.16 & 90.80 & 90.98 & 61.59 & 50.66 \\
\textsc{Crf}    & 91.24 & 90.92 & 91.08 & 61.92 & 50.33 \\
\textsc{Crf2o} & \textbf{91.56} & \textbf{91.11} & \textbf{91.33} & \textbf{63.08} & \textbf{50.99} \\[2pt]
\hline
\\[-9pt]
\multicolumn{6}{c}{CoNLL09} \\
\textsc{Loc}  & 79.20 & 79.02 & 79.11 & 40.10 & 28.91 \\
\textsc{Crf}    & 79.17 & 79.55 & 79.36 & 40.61 & 29.38 \\
\textsc{Crf2o} & \textbf{81.00} & \textbf{80.63} & \textbf{80.82} & \textbf{42.53} & \textbf{30.09} \\
\bottomrule
\end{tabular}
\caption{Sub- and full-tree performance on test data.}
\label{table:dev-test-subtree}

\end{table}


\paragraph{Impact of MBR decoding.} For \textsc{Crf} and \textsc{Crf2o}, we by default to perform MBR decoding, which employs the Eisner algorithm over marginal probabilities \cite{smith-smith-2007-probabilistic} to find the best tree.
\begin{equation}
\begin{split}
& {\boldsymbol{y}}^* = \arg\max_{\boldsymbol{y}} \left[\sum_{i \rightarrow j \in \boldsymbol{y}}{p(i \rightarrow j|\boldsymbol{x})} \right]
\end{split}
\end{equation}
Table~\ref{table:dev-test} reports the results of directly finding 1-best trees according to dependency scores. 
Except for PTB, probably due to the high accuracy already, MBR decoding brings small yet consistent improvements for both \textsc{Crf} and \textsc{Crf2o}.

\paragraph{Convergence behavior.}


Figure~\ref{fig:convergency} compares the convergence curves. 
For clarity, we plot one data point corresponding to the peak LAS every 20 epochs.
We can clearly see that both structural learning and high-order modeling
consistently improve the model.
\textsc{Crf2o} achieves steadily higher accuracy and converges much faster than the basic \textsc{Loc}.


\paragraph{Performance at sub- and full-tree levels.}

\begin{figure}[tb]
\centering
\subfigure{
    \begin{minipage}[b]{0.22\textwidth}
        \centering
        \includegraphics[width=1.05\textwidth]{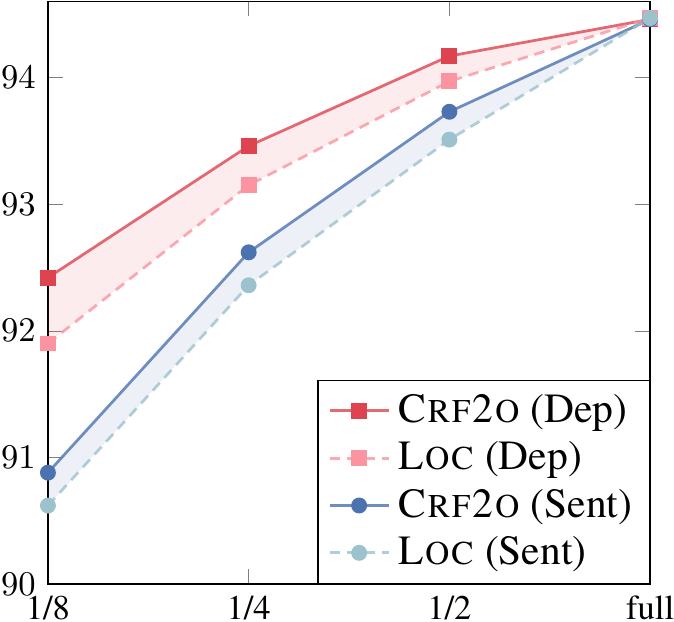}
    \end{minipage}
}
\subfigure{
    \begin{minipage}[b]{0.22\textwidth}
        \centering
        \includegraphics[width=1.05\textwidth]{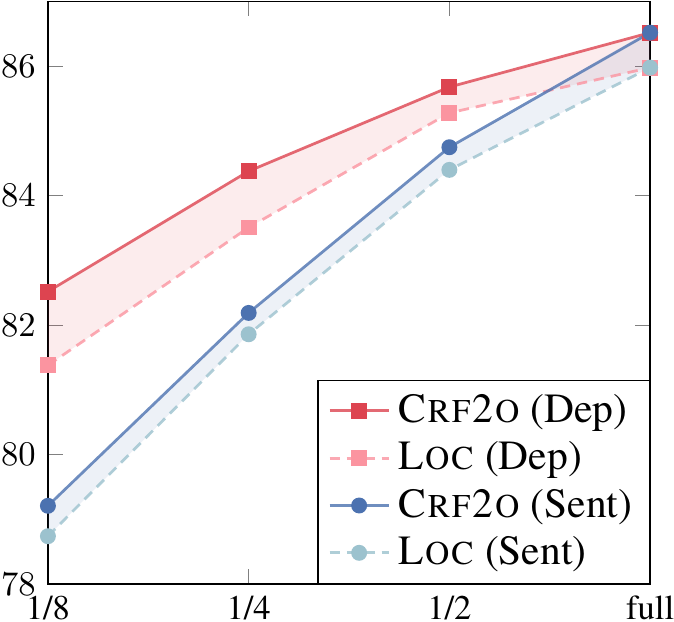}
    \end{minipage}
}
\caption{
 LAS on PTB (left) and CoNLL09-test (right) regarding the amount of training data (dependencies vs. sentences).
}
\label{fig:part-gap}
\end{figure}

\begin{table*}[tb]
\setlength{\tabcolsep}{3.4pt}
\centering
\begin{tabular}{lccccccccccccc}
\toprule
& bg & ca & cs & de & en & es & fr & it & nl & no & ro & ru & Avg.\\[1pt]
\hline
\\[-9pt]
\multicolumn{14}{c}{UD2.2} \\[1pt]
$\textsc{Loc}_{\textsc{mst}}$ &         90.44  &         91.11  &         91.04                   &         80.21                   &         86.86                   &         90.67                   &         87.99  &         91.19                    &         88.24                    &         90.35                   &         86.24                    &         93.01                    &         88.95 \\
\textsc{Loc}                  &         90.45  &         91.14  &         90.97                   &         80.02                   &         86.83                   &         90.56                   &         87.76  &         91.14                    &         87.72                    &         90.74                   &         86.20                    &         93.01                    &         88.88 \\
\textsc{Crf}                  &         90.73  &         91.25  &         91.01                   & \textbf{80.56}\rlap{$^\dagger$} &         86.92                   &         90.81\rlap{$^\dagger$}  & \textbf{88.16} &         91.64\rlap{$^\dagger$}   &         88.10                    &         90.85                   &         86.50                    &         93.17\rlap{$^\dagger$}   &         89.14\rlap{$^\ddagger$} \\
\textsc{Crf2o}                & \textbf{90.77} & \textbf{91.29} & \textbf{91.54}\rlap{$^\dagger$} &         80.46                   & \textbf{87.32}\rlap{$^\dagger$} & \textbf{90.86}\rlap{$^\dagger$} &         87.96  & \textbf{91.91}\rlap{$^\ddagger$} & \textbf{88.62}\rlap{$^\ddagger$} & \textbf{91.02}\rlap{$^\dagger$} & \textbf{86.90}\rlap{$^\ddagger$} & \textbf{93.33}\rlap{$^\ddagger$} & \textbf{89.33}\rlap{$^\ddagger$} \\[1pt]
\multicolumn{14}{c}{using raw text} \\[1pt]
Ji19           & 88.28 & 89.90 & 89.85 & 77.09 & 81.16 & 88.93 & 83.73 & 88.91 & 84.82 & 86.33 & 84.44 & 86.62 & 85.83 \\
\textsc{Crf2o} & \textbf{89.72} & \textbf{91.27} & \textbf{90.94}  & \textbf{78.26} & \textbf{82.88} & \textbf{90.79} & \textbf{86.33} & \textbf{91.02} & \textbf{87.92} & \textbf{90.17} & \textbf{85.71} & \textbf{92.49} & \textbf{88.13} \\
\hline
\\[-9pt]
\multicolumn{14}{c}{UD2.3} \\[1pt]
$\textsc{Loc}_{\textsc{mst}}$   &         90.56  &         91.03  &         91.98  &         81.59  & 86.83 &         90.64  & 88.23 &         91.67  &         88.20  &         90.63  & 86.51 & 93.03 &         89.23 \\
\textsc{Loc}                    &         90.57  &         91.10  &         91.85  &         81.68  & 86.54 &         90.47  & 88.40 &         91.53  &         88.18  &         90.65  & 86.31 & 92.91 &         89.19 \\
\textsc{Crf}   &         90.52  & \textbf{91.19} &         92.02  &         81.43  &         86.88\rlap{$^\dagger$}  &         90.76\rlap{$^\dagger$}  &         88.75  &         91.76  &         88.08  & \textbf{90.79} & 86.54 & 93.16\rlap{$^\ddagger$} &         89.32\rlap{$^\ddagger$} \\
\textsc{Crf2o} & \textbf{90.76} &         91.12  & \textbf{92.15}\rlap{$^\ddagger$} & \textbf{81.94} & \textbf{86.93}\rlap{$^\dagger$} & \textbf{90.81}\rlap{$^\ddagger$} &         \textbf{88.83}\rlap{$^\dagger$}  & \textbf{92.34}\rlap{$^\ddagger$} & \textbf{88.21}\rlap{$^\dagger$} & 90.78 & \textbf{86.62} & \textbf{93.22}\rlap{$^\ddagger$} & \textbf{89.48}\rlap{$^\ddagger$} \\
\multicolumn{14}{c}{using gold POS tags} \\[1pt]
Zhang19        & 90.15 & 91.39 & 91.10 & 83.39 & 88.52 & 90.84 & 88.59 & 92.49 & 88.37 & 92.82 & 84.89 & 93.11 & 89.85 \\
\textsc{Crf2o} & \textbf{91.32} & \textbf{92.57} & \textbf{92.66} & \textbf{84.56} & \textbf{88.98} & \textbf{91.88} & \textbf{89.83} & \textbf{92.94} & \textbf{89.85} & \textbf{93.26} & \textbf{87.39} & \textbf{93.86} & \textbf{90.76} \\
\bottomrule
\end{tabular}
\caption{LAS on UD2.2 and UD2.3 test datasets.
Again, $\dagger$ and $\ddagger$ means significance level at $p<0.05$ and $p<0.005$ respectively against the \textsc{Loc} parser. }
\label{table:ud2.3-test}
\end{table*}

Beyond the dependency-wise accuracy (UAS/LAS), we
would like to evaluate the models regarding performance at sub-tree and full-tree levels.
Table~\ref{table:dev-test-subtree} shows the results. We skip the partially labeled NLPCC19 data.
UCM means unlabeled complete matching rate, i.e., the percent of sentences obtaining whole correct skeletal trees, while
LCM further requires that all labels are also correct.

For SIB, we evaluate the model regarding unlabeled adjacent-sibling subtrees 
(system outputs vs. gold-standard references). 
According to Equation~\ref{eq:score-definition-2o},
$(i,k,j)$ is an adjacent-sibling subtree, if and only if $w_k$ and $w_j$ are both children of $w_i$ at the same side, and there are no other children of $w_i$ between them.
Given two trees, we can collect all adjacent-sibling subtrees and compose two sets of triples.
Then we evaluate the P/R/F values.
Please note that it is impossible to evaluate SIB for partially annotated references.

We can clearly see that by modeling adjacent-sibling subtree scores,
the SIB performance obtains larger improvement than both \textsc{Crf} and \textsc{Loc},
and this further contributes to the large improvement on full-tree matching rates (UCM/LCM).



\paragraph{Capability to learn from partial trees.}

To better understand why \textsc{Crf2o} performs very well on partially annotated NLPCC19,
we design more comparative experiments by retaining either a proportion of random training sentences (full trees) or a proportion of random dependencies for each sentence (partial trees).
Figure~\ref{fig:part-gap} shows the results.

We can see that the performance gap is quite steady
when we gradually reduce the number of training sentences.
In contrast, the gap clearly becomes larger when each training sentence has less annotated dependencies.
This shows that \textsc{Crf2o} is superior to the basic \textsc{Loc} in
utilizing partial annotated data for model training.




\subsection{Results on Universal Dependencies}

Table~\ref{table:ud2.3-test}
compares different models on UD datasets, which contain a lot of non-projective trees.
We adopt the pseudo-projective approach \cite{nivre-nilsson-2005-pseudo} for handling the ubiquitous non-projective trees of most languages. 
Basically, the idea is to transform non-projective trees into projective ones using more complex labels for post-processing recovery.



We can see that for the basic local parsers,
the direct non-projective $\textsc{Loc}_{\textsc{mst}}$ and the pseudo-projective \textsc{Loc}
achieve very similar performance.

More importantly, both \textsc{Crf} and \textsc{Crf2o} produce consistent improvements over the baseline in many languages.
On both UD2.2 and UD2.3, Our proposed \textsc{Crf2o} model achieves the highest accuracy for 10 languages among 12, and obtains significant improvement in more than 7 languages.
Overall, the averaged improvement is 0.45 and 0.29 on UD2.2 and UD2.3 respectively, which is also significant at $p<0.005$.

On average, our \textsc{Crf2o} parser outperforms \citet{ji-etal-2019-graph} by 2.30 on UD2.2 raw texts following CoNLL-2018 shared task setting, and \citet{zhang-etal-2019-empirical} by 0.91 on UD2.3 data with gold POS tags.
It is noteworthy that the German (de) result is kindly provided by Tao Ji after rerunning their parser with predicted XPOS tags, since their reported result in \citet{ji-etal-2019-graph} accidentally used gold-standard sentence segmentation, tokenization, and XPOS tags.
Our \textsc{Crf2o} parser achieves an average LAS of 87.64 using their XPOS tags.

\section{Related Works}
\label{section:relwork}


Batchification has been widely used in linear-chain CRF, but is rather complicated for tree structures.
\citet{eisner-2016-inside} presents a theoretical proof on the equivalence of outside and back-propagation for constituent tree parsing, and also briefly discusses other formalisms such as dependency grammar.
Unfortunately, we were unaware of Eisner's great work until we were surveying the literature for paper writing.
As an empirical study, we believe this work is valuable and makes it practical to deploy TreeCRF models in real-life systems.

\citet{falenska-kuhn-2019-non} present a nice analytical work on  dependency parsing, similar to \citet{gaddy-etal-2018-whats} on constituency parsing.
By extending the first-order graph-based parser of \citet{kiperwasser-goldberg-2016-simple}
into second-order, 
they try to find out
how much structural context is implicitly captured by the BiLSTM encoder.
They concatenate three BiLSTM output vectors ($i,k,j$) for scoring adjacent-sibling subtrees,
and adopt max-margin loss and the second-order Eisner decoding algorithm \cite{mcdonald-pereira-2006-online}.
Based on their negative results and analysis, they draw the conclusion that high-order modeling is redundant
because BiLSTM can implicitly and effectively encode enough structural context.
They also present a nice survey on the relationship between RNNs and syntax.
In this work, we use a much stronger basic parser 
and observe more significant UAS/LAS improvement than theirs.
Particularly, we present an in-depth analysis showing that explicitly high-order modeling
certainly helps the parsing model and thus is complementary to the BiLSTM encoder.

\citet{ji-etal-2019-graph} employ graph neural networks to incorporate high-order structural information into the biaffine parser implicitly.
They add a three-layer graph attention network (GAT) component \cite{velickovic2018graph} between the MLP and Biaffine layers.
The first GAT layer takes $\mathbf{r}_i^{h}$ and $\mathbf{r}_i^{m}$ from MLPs as inputs and produces new representation $\mathbf{r}_i^{h1}$ and $\mathbf{r}_i^{m1}$ by aggregating neighboring nodes. Similarly, the second GAT layer operates on $\mathbf{r}_i^{h1}$ and $\mathbf{r}_i^{m1}$, and produces $\mathbf{r}_i^{h2}$ and $\mathbf{r}_i^{m2}$.
In this way, a node gradually collects multi-hop  high-order information as global evidence for scoring single dependencies.
They follow the original local head-selection training loss.
In contrast, this work adopts global TreeCRF loss and explicitly incorporates high-order scores into the biaffine parser.

\citet{zhang-etal-2019-empirical} investigate the usefulness of structural training for the first-order biaffine parser.
They compare the performance of local head-selection loss, global max-margin loss, and TreeCRF loss on multilingual datasets.
They show that TreeCRF loss is overall slightly superior to max-margin loss, and LAS improvement from  structural learning is modest but significant for some languages.
They also show that structural learning (especially TreeCRF) substantially improves sentence-level complete matching rate, which is consistent with our findings.
Moreover, they explicitly compute the inside and outside algorithms on CPUs via Cython programming.
In contrast, this work proposes an efficient second-order TreeCRF extension to the biaffine parser,
and presents much more in-depth analysis to show the effect of both structural learning and high-order modeling.

\section{Conclusions}
\label{section:conclusions}

This paper for the first time presents second-order TreeCRF for neural dependency parsing using triaffine for explicitly scoring second-order subtrees.
We propose to batchify the inside algorithm to accommodate GPUs.
We also empirically verify that the complex outside algorithm can be implicitly performed via efficient back-propagation, which naturally produces gradients and
marginal probabilities.
We conduct experiments and detailed analysis on 27 datasets from 13 languages, and  find that structural learning and high-order modeling can further enhance  the state-of-the-art biaffine parser in various aspects: 1) better convergence behavior; 2) higher performance on sub- and full-tree levels; 3)
 better utilization of partially annotated data.






\section*{Acknowledgments}

The authors would like to thank: 1) the anonymous reviewers for the helpful comments, 2) Wenliang Chen for helpful discussions on high-order neural dependency parsing, 3) Tao Ji for kindly sharing the data and giving beneficial suggestions for the experiments on CoNLL18 datasets, 4)
Wei Jiang, Yahui Liu, Haoping Yang, Houquan Zhou and Mingyue Zhou for their help in paper writing and polishing.
This work was supported by National Natural Science Foundation of China (Grant No. 61876116, 61525205, 61936010) and a Project Funded by the Priority Academic Program Development (PAPD) of Jiangsu Higher Education Institutions.

\bibliography{acl2020}
\bibliographystyle{acl_natbib}

\appendix
\section{More on Efficiency}
\label{section:decoding-efficiency}
\paragraph{Training speed.}
During training, we greedily find the 1-best head for each word without tree constraints.
Therefore, the processing speed is faster than the evaluation phase. Specifically, for \textsc{Loc}, \textsc{Crf} and \textsc{Crf2o}, the average one-iteration training time is about 1min, 2.5min and 3.5min on PTB.
In other words, the parser consumes about 700/300/200 sentences per second.

\paragraph{MST decoding.}
As \citet{dozat-etal-2017-stanfords} pointed out, they adopted an ad-hoc approximate algorithm which does not guarantee to produce the highest-scoring tree, rather than the ChuLiu/Edmonds algorithm for MST decoding.
The time complexity of the ChuLiu/Edmonds algorithm is $O(n^2)$ under the optimized implementation of \citet{tarjan1977finding}.
Please see the discussion of \citet{mcdonald-etal-2005-non} for details.

For $\textsc{Loc}_{\textsc{mst}}$, we directly borrow the MST decoding approach in the original parser of \citet{Timothy-d17-biaffine}. $\textsc{Loc}_{\textsc{mst}}$ achieves 94.43 LAS on PTB-test (inferior to 94.47 of \textsc{Loc}, see Table~\ref{table:dev-test}), and its parsing speed is over 1000 sentences per second.

\paragraph{Faster decoding strategy.}
Inspired by the idea of ChuLiu/Edmonds algorithm, we can further improve the efficiency of the CRF parsing models by avoiding the Eisner decoding for some sentences.
The idea is that if by greedily assigning a local max-scoring head word to each word, we can already obtain a legal projective tree, then we omit the decoding process for the sentence.
We can judge whether an output is a legal tree (single root and no cycles) using the Tarjan algorithm in $O(n)$ time complexity.
Further, we can judge whether a tree is a projective tree also in a straightforward way very efficiently.
In fact, we find that more than 99\% sentences directly obtain legal projective trees on PTB-test by such greedy assignment on marginal probabilities first.
We only need to run the decoding algorithm for the left sentences.

\end{document}